\documentclass[10pt,twocolumn,letterpaper]{article}

\usepackage[pagenumbers]{cvpr} 

\definecolor{cvprblue}{rgb}{0.21,0.49,0.74}
\usepackage[pagebackref,breaklinks,colorlinks,allcolors=cvprblue]{hyperref}

\def\paperID{16} 
\def\confName{CVPR}
\def\confYear{2025}

\title{Sequence-Based Identification of First-Person Camera Wearers \\ in Third-Person Views}

\author{%
Ziwei Zhao$^1$ \quad Xizi Wang$^1$ \quad Yuchen Wang$^1$ \quad Feng Cheng$^2$ \quad David Crandall$^1$ \\ $^1$ Indiana University Bloomington \quad$^2$ByteDance \\ {\tt\small $^1$\{zz47,xiziwang,wang617,djcran\}@iu.edu \quad $^2$chengfeng2333@gmail.com
}
}
\usepackage{algorithmic}
\usepackage{algorithm}

\begin{document}
\maketitle
\begin{abstract}
The increasing popularity of egocentric cameras has generated growing interest in studying multi-camera interactions in shared environments. Although large-scale datasets such as Ego4D and Ego-Exo4D have propelled egocentric vision research, interactions between multiple camera wearers remain underexplored—a key gap for applications like immersive learning and collaborative robotics. To bridge this, we present TF2025, an expanded dataset with synchronized first- and third-person views. 
In addition, we introduce a sequence-based method to identify first-person wearers in third-person footage, combining motion cues and person re-identification. 
\end{abstract}    
\section{Introduction}
\label{sec:intro}

The growing popularity of wearable first-person (egocentric) cameras, along with rapid progress in virtual reality (VR) and augmented reality (AR) technologies, has spurred increasing interest in multi-camera interactions within the same environments. 
Egocentric cameras, often body-worn or head-mounted, are widely used in robotics, entertainment, and content creation, capturing dynamic, user-centric perspectives. In contrast, traditional third-person (exocentric) cameras provide a broader, contextual view, complementing first-person footage. As egocentric cameras gain popularity, their integration with exocentric views is increasingly important for unlocking a variety
of applications.

Datasets such as \textbf{Ego4D}~\cite{grauman2022ego4d} and \textbf{Ego-Exo4D}~\cite{grauman2024ego} have significantly advanced the field of egocentric vision. Ego4D provides a large-scale collection of first-person videos documenting everyday activities, serving as a foundational resource for egocentric vision research. Expanding on this, Ego-Exo4D introduces synchronized egocentric and exocentric views, creating a multi-perspective dataset. This enables novel studies into cross-view relationships, particularly for applications where first-person agents (e.g., robots or camera wearers) can benefit from widely available third-person video data to enhance their learning.

\begin{figure}[t]
    \centering
    \includegraphics[width=0.45\textwidth]{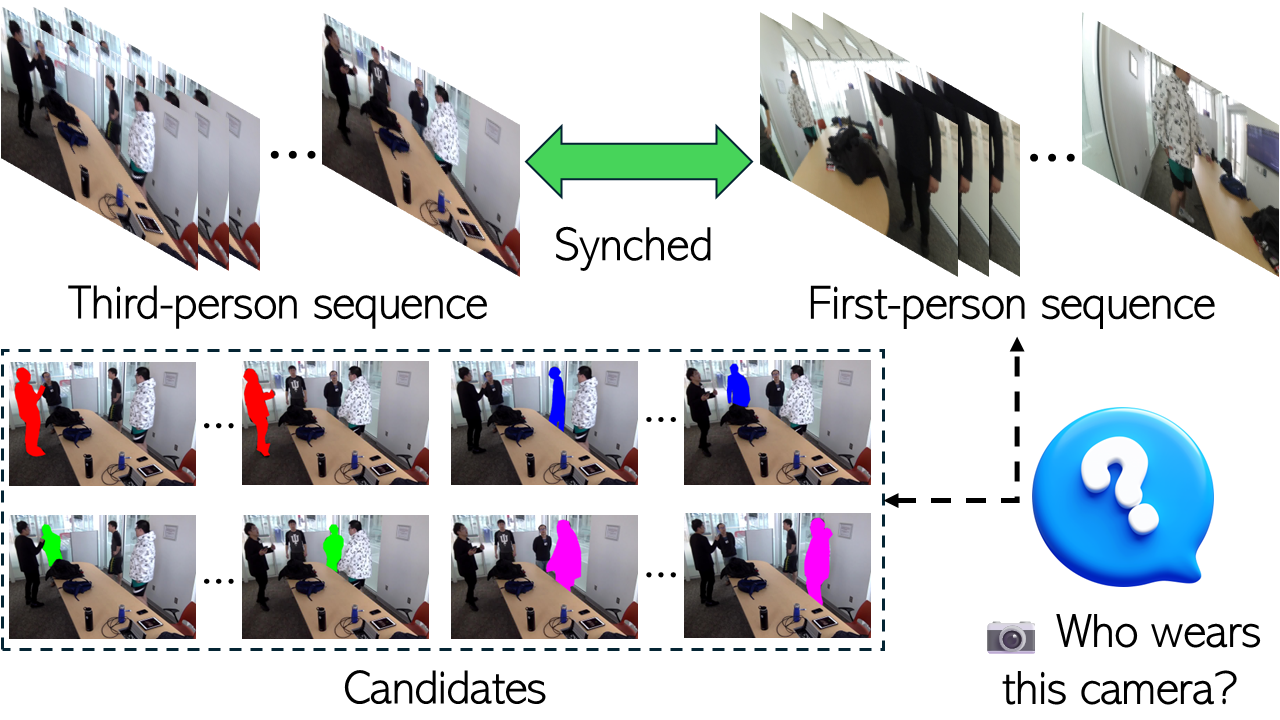}
    \caption{\textbf{Problem definition.}}
\end{figure}

However, Ego-Exo4D has a key limitation: it captures only one camera wearer per scenario, making it unsuitable for modeling interactions between multiple body-camera users. This prevents its applicability to real-world applications such as immersive learning in groups, multi-user VR experiences, and team-based robotics, where multiple egocentric cameras typically operate simultaneously. In such settings, identifying camera wearers in third-person views based on their first-person perspectives becomes essential for analyzing spatial relationships, role assignments, and coordinated actions within groups.

In this work, we introduce the \textbf{TF2025 (Third-First 2025)} dataset. TF2025 integrates videos from three sources: \textbf{TF2023}~\cite{zhao2024fusing}, \textbf{IUShareView}~\cite{xu2018joint}, and \textbf{Ego4D-TF}, a subset of \textbf{Ego4D}~\cite{grauman2022ego4d} enhanced with our additional annotated ground truth masks. TF2025 is 2.3 times larger than TF2023, the current largest dataset for this task, and includes more challenging frames to better reflect real-world scenarios. It also features three new train/test splits designed to evaluate multiple levels of generalizability, which could translate to different applications. In addition, we propose \textit{Motion Appearance Fusion (MAF)} for this task, a new framework that intergrates motion and appearance cues.
\section{Related Works}
Person identification has been explored in the context of first- and third-person cross-view understanding. Prior work has addressed this task using top-view cameras~\cite{ardeshir2016ego2top,ardeshir2018integrating,ardeshir2016egoreid}
, while our approach aligns with using side-view cameras as third-person perspectives~\cite{xu2018joint,fan2017identifying,zhao2024fusing}. Compared to top-view cameras, side-view cameras capture richer appearance details for camera wearers, at the expense of having a narrower field of view.

Recent progress in this domain includes multi-branch deep networks and semi-Siamese architectures that enhance cross-view alignment through joint embedding learning
~\cite{fan2017identifying,xu2018joint}. Notably, PEN~\cite{zhao2024fusing} advances this direction with a dual-branch framework that models personal relationships and environmental geometry. Our work extends upon these approaches by integrating temporal motion information with multi-cue information fusion, enabling more robust camera wearer identification in third-person views.
\section{Methodology}
\subsection{Problem Definition}

\begin{figure*}[h]
    \centering
    \includegraphics[width=0.66\textwidth]{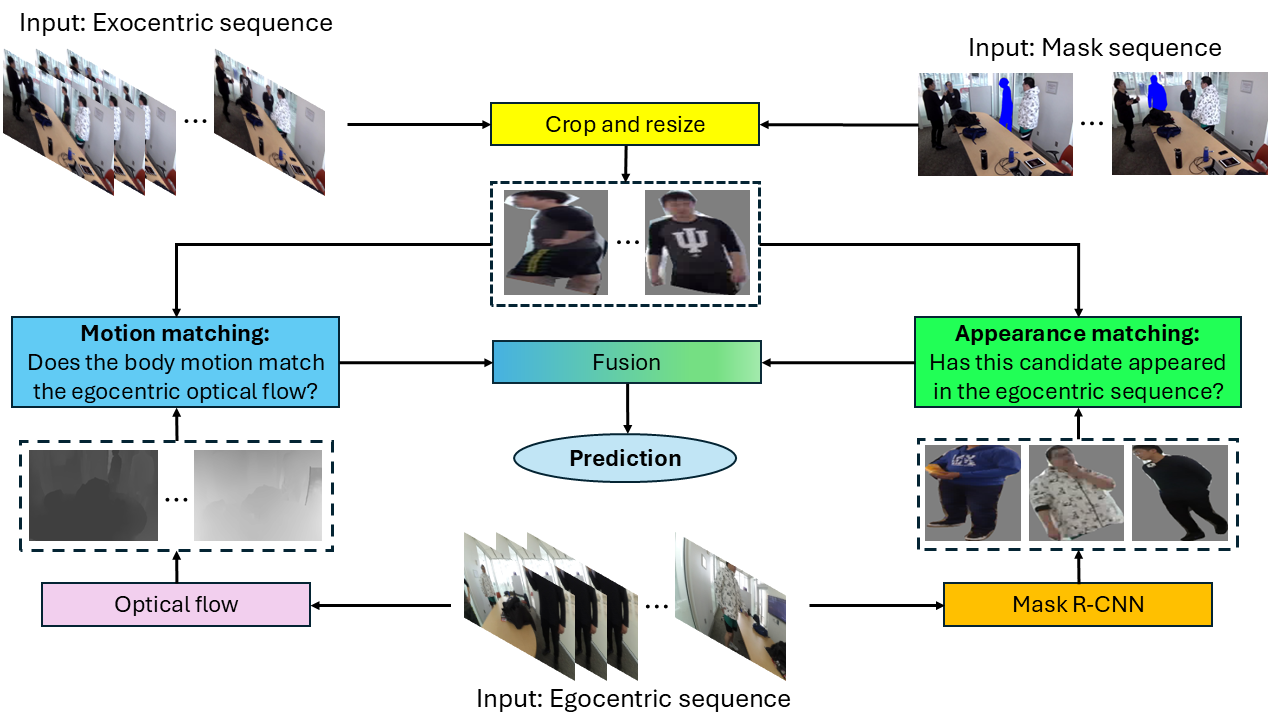}
    \caption{\textbf{Overview of the structure of our framework (MAF).} We utilize two models: \textit{motion matching}, which matches the motion of the candidate to the egocentric optical flow, and \textit{appearance matching} that checks if the candidate appeared in the first-person view. Then we integrates information from both sides to make the final prediction.}
    \label{fig:fullmodel}
\end{figure*}

We define the task of identifying camera wearer in third-person views as follows: Given a \( t \)-frame sequence of first-person frames \( Ego: \{Ego_1, Ego_2, \ldots, Ego_{\mathit{t}}\} \), a sequence of third-person frames \( Exo: \{Exo_1, Exo_2, \ldots, Exo_{\mathit{t}}\} \), and \( N \) mask sequences in the third-person view representing \( N \) candidate people,
\begin{align*}
\text{Candidate 1: } & \{Mask_{11}, Mask_{12}, \ldots, Mask_{1t}\} \\
\text{Candidate 2: } & \{Mask_{21}, Mask_{22}, \ldots, Mask_{2t}\} \\
& \vdots \\
\text{Candidate N: } & \{Mask_{N1}, Mask_{N2}, \ldots, Mask_{Nt}\},
\end{align*}
the goal is to identify the mask sequence which represents the correct candidate that corresponds to the first-person frame sequence \( Ego: \{Ego_1, Ego_2, \ldots, Ego_{\mathit{t}}\} \). 
\subsection{Motion Matching}
First, we build a model to link the body motion of camera wearers in the third-person view and the optical flow in the first-person view, denoted \textit{motion matching}.

We start by discussing the concept of motion within the context of this paper. Some egocentric datasets, such as \textbf{Ego4D} and \textbf{Ego-Exo4D}, utilize head-mounted and eyewear cameras, while others like \textbf{IUShareView} and \textbf{TF2023} use chest-worn cameras. Our approach does not restrict the prediction of motion to any predefined body part. Instead, we seek to predict movements from any body part that could induce changes (optical flow) in the first-person view. This design choice enhances the generalizability of our model, as body-worn cameras in real-world scenarios can be mounted on various body parts, such as the shoulder, wrist, or waist.

In addition, we consider two types of motion—\textbf{translation} and \textbf{rotation}—as they influence optical flow in first-person views in different ways. 
For \textbf{translational motion}, the camera's movement is represented by both the optical flow in pixel space and the depth of each pixel. This is because, during translation, objects closer to the camera exhibit faster apparent motion across the field of view compared to objects farther away, resulting in a depth-dependent optical flow pattern. In contrast, for \textbf{rotational motion}, the optical flow is depth-independent. The angular movement of the camera causes all points in the scene to shift uniformly in pixel space.

To model this, we apply a sequence-based backbone 
to a candidate in the third-person view, cropped using the third-person frames and the mask of the candidate. We then use a Multi-Layer Perceptron (MLP) to predict two values, $T_{exo}$ and $R_{exo}$, representing a combination of the total translational and rotational motion in the sequence.

For the first-person view, we compute two values, \( T_i \) and \( R_i \), that are related to the translational and rotational motion for frame $i$, 
\begin{equation}
T_{i} = \underset{j \in frame_i}{\mathrm{median}}\left(depth(p_{ij})\sqrt{fl_{x}(p_{ij})^{2} + fl_{y}(p_{ij})^{2}}\right)
\label{eq:T motion}
\end{equation}
\begin{equation}
R_{i} = \underset{j \in frame_i}{\mathrm{median}}\left(\sqrt{fl_{x}(p_{ij})^{2} + fl_{y}(p_{ij})^{2}}\right),
\label{eq:R motion}
\end{equation}
where \( p_{ij} \) denotes the \( j \)-th pixel in frame \( i \), and \( depth(p_{ij}) \), \( fl_{x}(p_{ij}) \), and \( fl_{y}(p_{ij}) \) represent the depth, x-axis optical flow, and y-axis optical flow, respectively. 
We then compute the total motion of the first-person view by aggregating the two motion values \( T_i \) and \( R_i \) across all frames, represented as a tuple of the cumulative translational and rotational motion, as shown in Eq.~\ref{eq:total motion},
\begin{equation}
    \mathrm{motion}(\{Ego_{1},\ldots,Ego_{t}\}) = \left(\sum_{i=1}^{t-1}T_{i}, \sum_{i=1}^{t-1}R_{i}\right).
\label{eq:total motion}
\end{equation}

During training, our motion matching model predicts two values, \( T_{exo} \) and \( R_{exo} \), for the input sequence. The loss function is computed as the sum of squared errors between the two predicted motion values and the corresponding translational motion and rotational motion ground truth values derived from the first-person sequence. At inference time, the candidate with the smallest sum of squared errors is selected as the prediction.

During inference, we employ a sliding window approach to evaluate motion similarity across overlapping sequences then calculate the sum of errors for translational and rotational motion within each window.

\subsection{Appearance Matching}
The presence of individuals in the first-person view provides additional cues for matching, as camera wearers should not see themselves in their own first-person views. For this purpose, we apply a person re-ID model, denoted \textit{appearance matching}. During training, the backbone outputs a feature vector, and an attached MLP head generates classification labels, optimized using cross-entropy loss for classification and triplet loss for metric learning. For inference, we compute the L2 distance between pairs of individuals using their feature vectors.

During inference, we select 3 frames (first, middle, end) from the first-person sequence and apply \textit{Mask R-CNN} to crop out all detected individuals. Each detected individual and third-person candidates are passed through the same backbone, and we compute its average L2 distance from the third-person candidate across all frames. These three frames were chosen as a balance between capturing most individuals appearing in the first-person sequence and avoiding excessive detections of the same person, which could complicate the fusion process.

\subsection{Confidence-Based Adaptive Fusing}
After \textit{motion matching} and \textit{appearance matching}, we have the following results for \( N \) candidates in third-person and \( M \) detected individuals in the first-person sequence:
\begin{align*}
\text{Motion scores: } & \{Motion_{1}, \ldots, Motion_{N}\} \\
\text{Appear. scores 1: } & \{Appearance_{11}, \ldots, Appearance_{1N}\} \\
& \vdots \\
\text{Appear. scores M: } & \{Appearance_{M1}, \ldots, Appearance_{MN}\}
\end{align*}

The goal of this task is to identify the correct candidate among \(\{1, \dots, N\}\) that matches the first-person sequence. This poses several challenges for the fusion method:
\begin{enumerate}
    \item Both \( M \) and \( N \) are different for each query.
    \item The same individual can be detected multiple times in the 3 chosen frames.
    \item Motion scores and appearance scores behave differently. A lower motion score reflects better matching between the third/first-person motion, while a lower appearance score means the candidate is seen in the first-person view, so unlikely to be the camera wearer.
\end{enumerate}

We propose \textit{Confidence-Based Adaptive Fusing (CBAF)}, a novel module designed to integrate results from the two preceding modules. \textit{CBAF} adaptively selects which source of information to use for candidate elimination or final prediction, requires no training for different datasets, and can handle arbitrary values of \( M \) and \( N \).

We first define the concept of confidence in the context of this paper,
\[
\text{Confidence} = \lambda_{\text{Appearance}} \cdot \alpha_{\text{mask}} \cdot \frac{x_{(2)}}{x_{(1)}},
\]
where:
\begin{itemize}
    \item \( x_{(1)} \) is the smallest value among \( N \) values from the same source,
    \item \( x_{(2)} \) is the second smallest value among \( N \) values from the same source,
    \item \( \lambda_{\text{Appearance}} \) is a constant representing our trust in the \textit{appearance matching} model, which depends on prior knowledge about the dataset (e.g., how clearly individuals are visible). For motion scores, \( \lambda_{\text{Appearance}} = 1 \), and
    \item \( \alpha_{\text{mask}} \) is the confidence from \textit{Mask R-CNN} when detecting this individual. For motion scores, \( \alpha_{\text{mask}} = 1 \).
\end{itemize}

Essentially, this calculates the ratio between the two smallest items from each score source. A higher confidence value indicates a clearer distinction between the most prominent candidate and the runner-up, while a lower confidence value suggests ambiguity between the two candidates. This ratio helps us evaluate the reliability of each source of information, guiding how much trust we place in their respective results.

The \textit{CBAF} algorithm uses the following logic: we compare the confidence of motion scores against that of all appearance confidence scores. If the motion scores exhibit the highest confidence, the candidate with the lowest motion score is selected as the prediction. Conversely, if an appearance source yields the highest confidence, the corresponding candidate is removed from all sources, and that source is subsequently eliminated.

\section{Dataset }
We introduce the \textbf{TF2025} dataset. TF2025 uses videos from three sources: TF2023~\cite{zhao2024fusing}, IUShareView~\cite{xu2018joint} and Ego4d~\cite{grauman2022ego4d}. For every frame in the third-person views, we created segmentation masks for all actors and assigned IDs to link the masks with their corresponding first-person camera wearers. \textbf{TF2023} is currently the largest available dataset for this task. 
In contrast to TF2023, which focused on frames with active social interactions, TF2025 utilizes all source frames from the videos of TF2023, including more challenging egocentric sequences with minimal visual context (e.g., looking at ceiling or whiteboards).

\textbf{IUShareView} is another dataset recorded under the same settings as TF2023, using the same camera models and overlapping actors.

For \textbf{Ego4D}, we annotated 5 board game scenes featuring $\geq$3 camera wearers. For each scene, we designated one camera wearer as pseudo-third-person view, while using the remaining camera wearers as first-person views. This annotated subset, \textbf{Ego4D-TF}, enables cross-dataset evaluation of different camera configurations (head- vs.\ chest-mounted) and different activities for camera wearers.

TF2025 is $2.3\times$ larger than TF2023, the current largest dataset available for this task. In addition, we introduce three new train/test splits for TF2025:
\begin{itemize}
    \item Seen: The first 80\% of each video from TF2023 and IUShareView are used for training, and the remaining 20\% for testing.
    \item Unseen: TF2023 and IUShareView are divided at video level such that the same camera wearer do not appear in both training and testing.
    \item Cross-dataset: We use TF2023 and IUShareView for training and Ego4d-TF for testing.
\end{itemize}

\section{Conclusion}
This extended abstract presents ongoing research on identifying first-person camera wearers in third-person views. We introduce \textbf{TF2025}, an expanded dataset based on IUShareView, TF2023, and Ego4d. TF2025 is larger than existing datasets for this task while also includes multiple train-test splits for evaluation to evaluate different levels of generalizability. In addition, we propose \textit{Motion Appearance Fusion}, a novel framework that integrates information from motion matching and person re-identification. Future work includes adding visualizations for the TF2025 dataset and conducting more experiments on the model's design.

\section{Acknowledgments}
This
work was supported in part by the National Science Foundation under award DRL-2112635 to the AI Institute for Engaged Learning. Any opinions, findings, and conclusions
or recommendations expressed in this material are those of
the author(s) and do not necessarily reflect the views of the
National Science Foundation.
{
    \bibliographystyle{ieeenat_fullname}
    \bibliography{main}
}


\end{document}